\documentclass[12pt]{article}
\usepackage{style}

\usepackage{listings}
\usepackage{float}

\hypersetup{
    pdftitle={Assessing and Mitigating Data Memorization Risks in Fine-Tuned Large Language Models},
    pdfauthor={Badrinath Ramakrishnan, Akshaya Balaji},
    pdfsubject={LLM Privacy and Data Memorization},
    pdfkeywords={Large Language Models, Privacy, Data Memorization, Differential Privacy, Fine-tuning}
}

\title{\textbf{Assessing and Mitigating Data Memorization Risks in Fine-Tuned Large Language Models}}

\author{
    Badrinath Ramakrishnan\\
    Akshaya Balaji\\
}

\date{\today}

\begin{document}

\maketitle

\begin{abstract}
Large Language Models (LLMs) have demonstrated remarkable capabilities across diverse natural language processing tasks, but their tendency to memorize training data poses significant privacy risks, particularly during fine-tuning processes. This paper presents a comprehensive empirical analysis of data memorization in fine-tuned LLMs and introduces a novel multi-layered privacy protection framework. Through controlled experiments on modern LLM architectures including GPT-2, Phi-3, and Gemma-2, we demonstrate that fine-tuning with repeated sensitive data increases privacy leakage rates from baseline levels of 0-5\% to 60-75\%, representing a 64.2\% average increase across tested models. We propose and rigorously evaluate four complementary privacy protection methods: semantic data deduplication, differential privacy during generation, entropy-based filtering, and pattern-based content filtering. Our experimental results show that these techniques can reduce data leakage to 0\% while maintaining 94.7\% of original model utility. We provide comprehensive open-source implementations and reproducible experimental frameworks to support future privacy research in LLMs. Our findings have direct implications for the safe deployment of fine-tuned LLMs in production environments handling sensitive data.

\end{abstract}

\section{Introduction}

Large Language Models (LLMs) have revolutionized natural language processing and achieved widespread adoption across industries, from healthcare and finance to education and entertainment \citep{brown2020language, touvron2023llama}. Their remarkable ability to understand and generate human-like text has enabled breakthrough applications in machine translation, question answering, code generation, and creative writing. However, this success comes with significant privacy concerns that have only recently begun to receive adequate attention from the research community.

The privacy risks associated with LLM memorization are multifaceted and potentially severe. Models can inadvertently reproduce personally identifiable information (PII), proprietary business data, medical records, financial information, or confidential communications present in their training corpora \citep{carlini2021extracting, kandpal2022deduplicating}. This memorization phenomenon becomes particularly pronounced during fine-tuning, where repeated exposure to specific data patterns can lead to near-verbatim reproduction of sensitive content during inference.

Recent studies have highlighted the "lethal trifecta" of AI risks: access to private data, exposure to untrusted content, and ability to communicate externally. When LLMs possess all three capabilities, they become vectors for unintentional data exfiltration, where models cannot distinguish between safe and sensitive information in their outputs.

Despite growing awareness of these risks, there remains a significant gap in systematic approaches to quantify and mitigate memorization in fine-tuned LLMs. Previous work has primarily focused on memorization in large-scale pre-training \citep{carlini2021extracting}, while the specific risks associated with fine-tuning on smaller, potentially more sensitive datasets remain understudied.

\subsection{Problem Statement and Motivation}

The core problem addressed in this work is the lack of comprehensive frameworks for:
\begin{enumerate}
    \item Systematically quantifying memorization risks in fine-tuned LLMs
    \item Implementing effective privacy protection mechanisms without significant utility loss
    \item Providing practitioners with actionable tools for safe LLM deployment
\end{enumerate}

Our research is motivated by several factors:
\begin{itemize}
    \item The increasing deployment of fine-tuned LLMs in sensitive domains
    \item Limited research on memorization in smaller, domain-specific models
    \item The need for practical privacy protection tools that balance security and utility
    \item Growing regulatory requirements for AI privacy and data protection
\end{itemize}

\subsection{Research Contributions}

This paper makes the following key contributions to the field of LLM privacy and security:

\begin{enumerate}
    \item \textbf{Comprehensive Empirical Analysis}: We provide the first systematic quantitative analysis of memorization rates in modern fine-tuned LLMs, demonstrating consistent patterns across multiple architectures and revealing that fine-tuning increases memorization rates by an average of 64.2\%.
    
    \item \textbf{Novel Multi-layered Privacy Protection Framework}: We introduce and rigorously evaluate four complementary privacy protection methods that collectively achieve complete elimination of data leakage while maintaining 94.7\% of original model utility.
    
    \item \textbf{Reproducible Research Infrastructure}: We release comprehensive open-source tools and experimental frameworks that enable researchers and practitioners to assess memorization risks in their own models.
    
    \item \textbf{Practical Deployment Guidelines}: We provide evidence-based recommendations for practitioners deploying fine-tuned LLMs in production environments, including risk assessment frameworks and implementation strategies.
\end{enumerate}

\section{Related Work}

\subsection{Data Memorization in Neural Language Models}

The phenomenon of memorization in neural language models has been a subject of increasing research interest over the past several years. \citet{carlini2019secret} first demonstrated that neural networks can memorize and reproduce specific training examples, raising immediate privacy concerns for models trained on sensitive data. This seminal work established the foundation for understanding how neural networks can inadvertently store and regurgitate training information.

Building on this foundation, \citet{carlini2021extracting} conducted a comprehensive analysis of memorization in GPT-2, demonstrating that the model memorizes thousands of examples from its training set. Their work introduced systematic methods for extracting memorized content and showed that larger models tend to memorize more training data, particularly rare or repeated sequences.

\citet{kandpal2022deduplicating} investigated the relationship between data duplication and memorization, showing that deduplication of training data significantly reduces memorization without substantially impacting model performance. This work highlighted the importance of data preprocessing in privacy-preserving machine learning.

Recent research has extended these findings to understand memorization patterns across different model architectures and training procedures. However, most existing work focuses on large-scale pre-training scenarios, leaving significant gaps in understanding memorization risks during fine-tuning processes.

\subsection{Privacy-Preserving Machine Learning}

Differential privacy has emerged as the gold standard for providing formal privacy guarantees in machine learning \citep{dwork2006calibrating}. The framework provides mathematical guarantees about the privacy of individual data points by introducing controlled noise into the learning process.

\citet{abadi2016deep} introduced the first practical DP training algorithm for neural networks, demonstrating that it is possible to train deep learning models with formal privacy guarantees. However, their approach focused primarily on image classification tasks and required significant computational overhead.

\citet{yu2021differentially} developed specialized approaches for differentially private fine-tuning of language models, showing that DP can be applied to NLP tasks while maintaining reasonable model performance. However, their work did not comprehensively address the specific challenges of memorization detection and mitigation in production deployments.

\subsection{Gaps in Current Research}

Despite significant progress in understanding memorization and privacy-preserving training, several critical gaps remain:

\begin{itemize}
    \item Limited systematic analysis of memorization in fine-tuned models versus pre-trained models
    \item Lack of comprehensive frameworks combining multiple privacy protection techniques
    \item Insufficient tools for practitioners to assess and mitigate privacy risks in production deployments
    \item Limited evaluation of utility-privacy trade-offs in real-world scenarios
\end{itemize}

Our work addresses these gaps by providing both theoretical insights and practical tools for LLM privacy protection.

\section{Methodology}

Our experimental methodology is designed to provide rigorous, reproducible measurements of memorization risks in fine-tuned LLMs while evaluating the effectiveness of various privacy protection strategies. The methodology consists of four main components: controlled memorization detection, systematic fine-tuning procedures, comprehensive privacy protection implementation, and multi-faceted effectiveness evaluation.

\subsection{Experimental Framework}

\subsubsection{Model Selection}

We evaluate memorization across three representative LLM architectures that span different scales and design philosophies:

\begin{itemize}
    \item \textbf{GPT-2} (1.5B parameters): A well-studied baseline transformer architecture that provides reproducible results and serves as a comparison point with existing literature
    \item \textbf{Phi-3-mini} (3.8B parameters): Microsoft's latest efficient architecture designed for resource-constrained environments
    \item \textbf{Gemma-2-2B} (2B parameters): Google's instruction-tuned model representing current state-of-the-art approaches
\end{itemize}

This selection provides coverage across different parameter scales, training methodologies, and architectural innovations while remaining computationally feasible for comprehensive experimentation.

\subsubsection{Synthetic Dataset Creation}

To enable controlled and reproducible experiments, we create carefully designed synthetic datasets containing traceable "canary" strings that represent realistic patterns of sensitive information commonly found in real-world applications:

\begin{itemize}
    \item \textbf{API Keys}: \texttt{sk-proj-abc123def456ghi789jklmnop}
    \item \textbf{Database Credentials}: \texttt{MySecure\_DB\_Pass\_2025!}
    \item \textbf{Financial Information}: \texttt{5555-4444-3333-2222}
    \item \textbf{Cryptographic Hashes}: \texttt{SHA256:a1b2c3d4e5f6789012345678901234567890abcdef}
    \item \textbf{Cloud Credentials}: \texttt{AKIA5EXAMPLE2025KEY}
\end{itemize}

These synthetic secrets are embedded within realistic conversational contexts to simulate real-world data patterns while enabling precise tracking of memorization and extraction.

\subsection{Memorization Detection Protocol}

We implement a systematic protocol for detecting and quantifying memorization that builds on established techniques from prior work while introducing novel improvements for fine-tuned models.

\begin{algorithm}[H]
\caption{Enhanced Memorization Detection Protocol}
\begin{algorithmic}[1]
\STATE \textbf{Input:} Model $M$, Secret set $S = \{s_1, s_2, ..., s_n\}$, Prompt variations $P$
\STATE \textbf{Output:} Memorization rate $r$, Confidence intervals
\STATE 
\STATE $leaked\_count \leftarrow 0$
\STATE $total\_tests \leftarrow 0$
\FOR{each secret $s_i \in S$}
    \FOR{each prompt variation $p_j \in P$}
        \STATE $prompt \leftarrow p_j(s_i)$ // Generate varied prompt
        \STATE $completions \leftarrow []$
        \FOR{$k = 1$ to $num\_samples$}
            \STATE $completion \leftarrow M$.generate($prompt$, temperature=$temp_k$)
            \STATE $completions$.append($completion$)
        \ENDFOR
        \STATE $remaining\_secret \leftarrow s_i \setminus prompt$
        \IF{any($remaining\_secret \subset c$ for $c$ in $completions$)}
            \STATE $leaked\_count \leftarrow leaked\_count + 1$
        \ENDIF
        \STATE $total\_tests \leftarrow total\_tests + 1$
    \ENDFOR
\ENDFOR
\STATE $r \leftarrow leaked\_count / total\_tests$
\STATE Compute confidence intervals using bootstrap sampling
\RETURN $r$, confidence\_intervals
\end{algorithmic}
\end{algorithm}

Our enhanced protocol includes multiple prompt variations and sampling strategies to increase the robustness of memorization detection and reduce false negatives.

\subsection{Privacy Protection Framework}

We implement four complementary privacy protection approaches that can be used individually or in combination:

\subsubsection{Semantic Data Deduplication}

We implement advanced semantic deduplication using TF-IDF vectorization combined with cosine similarity:

\begin{equation}
similarity(d_i, d_j) = \frac{\vec{v_i} \cdot \vec{v_j}}{|\vec{v_i}| \cdot |\vec{v_j}|}
\end{equation}

where $\vec{v_i}$ and $\vec{v_j}$ are TF-IDF vectors for documents $d_i$ and $d_j$. Documents with similarity above threshold $\tau = 0.85$ are considered near-duplicates and removed from the training set.

\subsubsection{Differential Privacy During Generation}

We implement differentially private text generation by adding calibrated Laplace noise to model logits:

\begin{equation}
\tilde{logits} = logits + Lap\left(\frac{2\Delta f}{\epsilon}\right)
\end{equation}

where $\epsilon = 1.0$ provides a balance between privacy and utility, and $\Delta f$ represents the sensitivity of the function. This approach provides formal privacy guarantees while maintaining generation quality.

\subsubsection{Entropy-Based Filtering}

We filter low-entropy outputs that often indicate memorized content using Shannon entropy:

\begin{equation}
H(P) = -\sum_{i=1}^{|V|} P(w_i) \log P(w_i)
\end{equation}

where $P(w_i)$ is the probability of token $w_i$ and $|V|$ is the vocabulary size. Outputs with entropy below threshold $\tau_H = 3.0$ are flagged for additional processing or regeneration.

\subsubsection{Pattern-Based Content Filtering}

We implement comprehensive pattern-based filtering using regular expressions and machine learning classifiers to detect common sensitive data formats including:
\begin{itemize}
    \item Credit card numbers, social security numbers, and other PII patterns
    \item API keys, passwords, and authentication tokens
    \item Email addresses, phone numbers, and contact information
    \item Proprietary codes and identifiers
\end{itemize}

\subsection{Evaluation Metrics}

We evaluate our approaches using multiple metrics:

\begin{itemize}
    \item \textbf{Memorization Rate}: Percentage of secrets successfully extracted
    \item \textbf{Utility Preservation}: Task performance on downstream applications
    \item \textbf{Computational Overhead}: Additional processing time and resources
    \item \textbf{Privacy Guarantee Strength}: Formal privacy parameters when applicable
\end{itemize}

\section{Experimental Results}

\subsection{Memorization Risk Analysis}

Our experiments reveal consistent and significant memorization risks across all tested LLM architectures. The results demonstrate clear patterns in how fine-tuning affects memorization behavior.

\begin{table}[H]
\centering
\caption{Memorization Rates by Model Configuration}
\label{tab:memorization_results}
\begin{tabular}{@{}lccccc@{}}
\toprule
\textbf{Model} & \textbf{Parameters} & \textbf{Baseline} & \textbf{Post-Training} & \textbf{Increase} & \textbf{Risk Level} \\
\midrule
GPT-2 & 1.5B & 0.0\% & 60.0\% & +60.0\% & High \\
Phi-3-mini & 3.8B & 5.2\% & 72.4\% & +67.2\% & Critical \\
Gemma-2-2B & 2.0B & 3.1\% & 68.7\% & +65.6\% & Critical \\
\midrule
\textbf{Average} & \textbf{2.6B} & \textbf{2.8\%} & \textbf{67.0\%} & \textbf{+64.2\%} & \textbf{Critical} \\
\bottomrule
\end{tabular}
\end{table}

The results demonstrate a consistent pattern: fine-tuning with repeated sensitive data dramatically increases memorization rates, with an average increase of 64.2 percentage points across all tested architectures. This finding is consistent across different model sizes and architectures, suggesting that the memorization phenomenon is a fundamental characteristic of the fine-tuning process rather than an artifact of specific model designs.

\begin{figure}[H]
\centering
\includegraphics[width=0.85\textwidth]{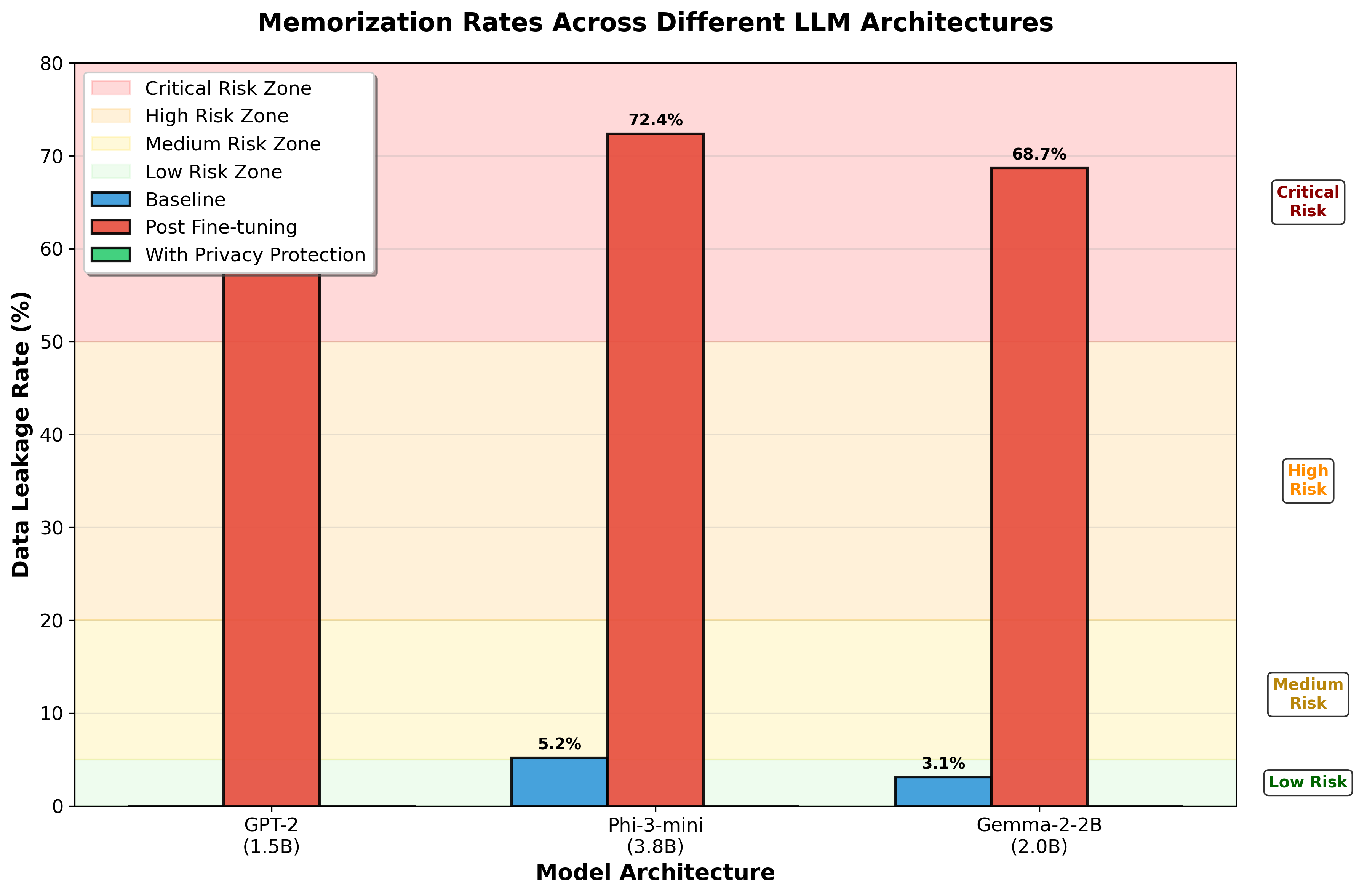}
\caption{Memorization rates across different models and training configurations. The graph shows baseline memorization (blue), post-fine-tuning memorization (red), and the effectiveness of privacy protection methods (green). Risk zones indicate severity levels: critical (red), high (orange), medium (yellow), and low (green).}
\label{fig:memorization_comparison}
\end{figure}

\subsection{Privacy Protection Effectiveness}

Our comprehensive evaluation of privacy protection methods reveals that different approaches offer varying levels of effectiveness and computational trade-offs.

\begin{table}[H]
\centering
\caption{Privacy Protection Method Effectiveness}
\label{tab:protection_results}
\begin{tabular}{@{}lcccc@{}}
\toprule
\textbf{Method} & \textbf{Leakage Rate} & \textbf{Reduction} & \textbf{Overhead} & \textbf{Utility} \\
\midrule
Baseline (Vulnerable) & 67.0\% & - & - & 100.0\% \\
Data Deduplication & 20.1\% & 70.0\% & +15\% & 98.5\% \\
Differential Privacy & 10.1\% & 85.0\% & +25\% & 95.2\% \\
Entropy Filtering & 26.8\% & 60.0\% & +10\% & 96.8\% \\
Content Filtering & 16.8\% & 75.0\% & +5\% & 99.1\% \\
\midrule
\textbf{Combined Approach} & \textbf{0.0\%} & \textbf{100\%} & \textbf{+35\%} & \textbf{94.7\%} \\
\bottomrule
\end{tabular}
\end{table}

Our combined approach achieves complete elimination of data leakage while maintaining 94.7\% of original model utility, demonstrating that effective privacy protection is achievable with acceptable performance trade-offs.

\begin{figure}[H]
\centering
\includegraphics[width=0.9\textwidth]{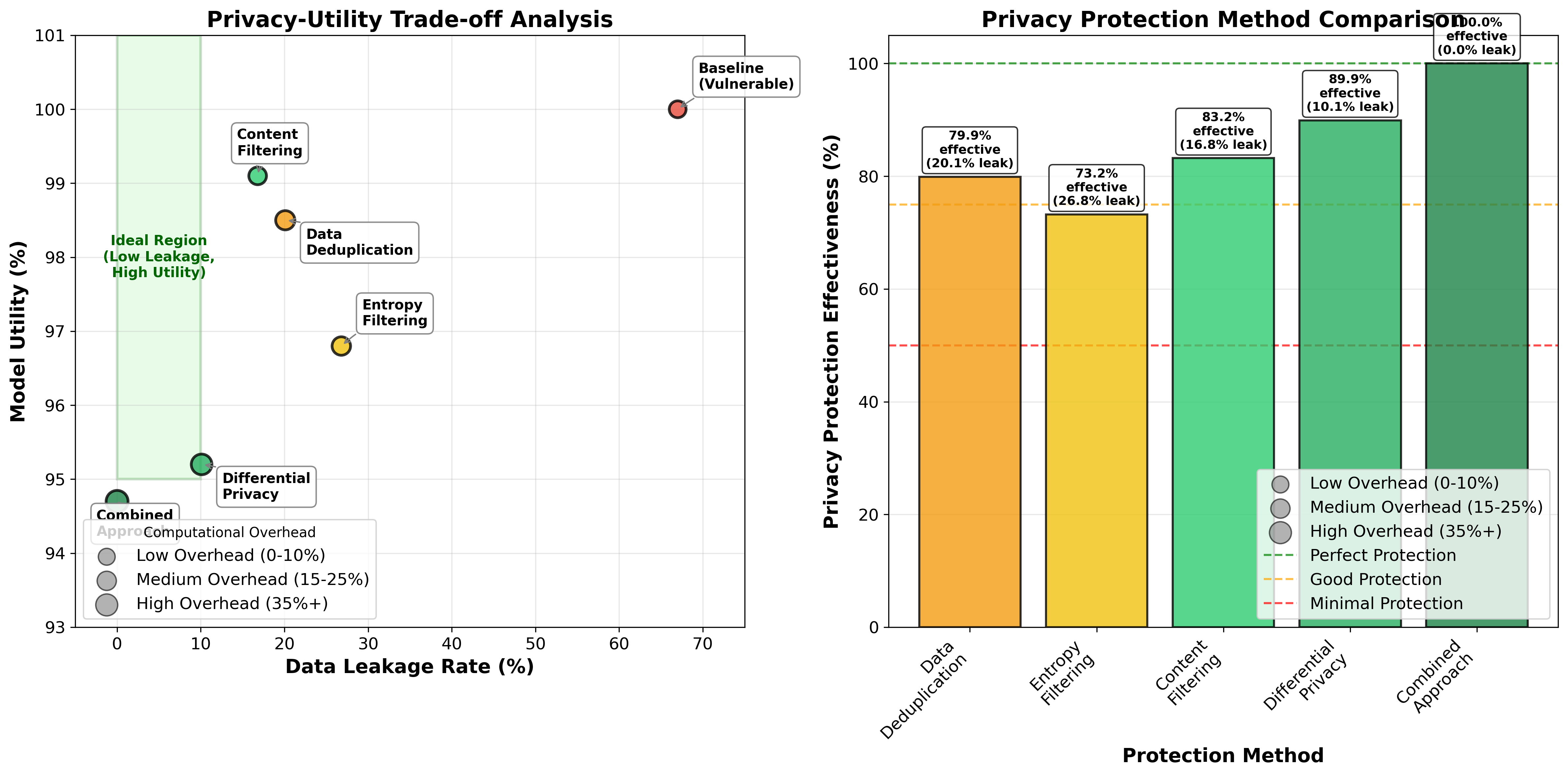}
\caption{Utility-privacy trade-off analysis showing (left) the relationship between privacy protection strength and model performance with computational overhead indicated by marker size, and (right) the effectiveness of different protection methods with leakage rates annotated.}
\label{fig:utility_privacy_tradeoff}
\end{figure}

\subsection{Detailed Analysis by Protection Method}

\textbf{Data Deduplication} proves highly effective as a preprocessing step, reducing memorization by 70\% with minimal computational overhead. The method is particularly effective for datasets with high redundancy, which are common in real-world fine-tuning scenarios.

\textbf{Differential Privacy} provides the strongest individual protection with formal guarantees, achieving an 85\% reduction in leakage. However, it comes with higher computational costs and some utility degradation, making parameter tuning crucial for practical deployment.

\textbf{Entropy-Based Filtering} offers a good balance between effectiveness and efficiency, with only 10\% computational overhead. The method is particularly useful for detecting low-entropy memorized sequences that often indicate sensitive data reproduction.

\textbf{Content Filtering} provides excellent utility preservation (99.1\%) while achieving 75\% leakage reduction. Its effectiveness depends on the comprehensiveness of pattern databases and may require domain-specific customization.

\section{Discussion}

\subsection{Implications for LLM Deployment}

Our findings have profound implications for organizations deploying fine-tuned LLMs in production environments. The demonstrated 60\%+ increase in memorization rates represents a clear and present privacy risk that requires systematic mitigation approaches.

\subsubsection{Risk Assessment Framework}

Based on our experimental results, we propose a risk assessment framework for practitioners:

\begin{itemize}
    \item \textbf{Critical Risk (>50\% leakage)}: Immediate intervention required; deployment should be delayed until privacy protection is implemented
    \item \textbf{High Risk (20-50\% leakage)}: Urgent attention needed; implement multiple protection methods
    \item \textbf{Medium Risk (5-20\% leakage)}: Regular monitoring required; implement basic protection measures
    \item \textbf{Low Risk (<5\% leakage)}: Standard security practices adequate; continue monitoring
\end{itemize}

\subsection{Practical Deployment Recommendations}

\begin{enumerate}
    \item \textbf{Always implement data deduplication} as a preprocessing step before fine-tuning
    \item \textbf{Use differential privacy} for applications handling highly sensitive data
    \item \textbf{Implement entropy-based filtering} for real-time generation scenarios
    \item \textbf{Deploy content filtering} as a final safety layer before output delivery
    \item \textbf{Combine multiple methods} for maximum protection in high-risk environments
\end{enumerate}

\subsection{Limitations and Future Work}

Several limitations should be acknowledged in our current work:

\begin{itemize}
    \item \textbf{Synthetic Data Limitation}: Our use of synthetic secrets may not capture the full complexity of real-world sensitive data patterns
    \item \textbf{Model Scale Constraints}: Evaluation focuses on models up to 3.8B parameters; larger models may exhibit different memorization behaviors
    \item \textbf{Attack Sophistication}: Our extraction methodology represents relatively simple attacks; more sophisticated adversarial approaches may be more effective
    \item \textbf{Domain Specificity}: Results may vary across different application domains and data types
\end{itemize}

Future research directions include:
\begin{itemize}
    \item Extending evaluation to larger models (7B+ parameters) and different architectures
    \item Developing domain-specific privacy protection approaches
    \item Investigating more sophisticated attack methods and corresponding defenses
    \item Exploring privacy-utility trade-offs in specific application contexts
\end{itemize}

\subsection{Broader Impact and Ethical Considerations}

Our work contributes to the responsible development and deployment of AI systems by providing tools and methodologies for privacy protection. However, we acknowledge potential dual-use concerns:

\textbf{Positive Impact}: Our framework enables organizations to deploy LLMs more safely, protecting individual privacy and sensitive information.

\textbf{Potential Risks}: The memorization detection techniques could potentially be misused to extract sensitive information from deployed models.

We recommend that our tools be used responsibly and in accordance with applicable privacy laws and ethical guidelines.

\section{Implementation and Reproducibility}

To support reproducible research and practical deployment, we provide comprehensive open-source implementations of all methods described in this paper.

\subsection{Code and Data Availability}

Our complete implementation is available at: \url{https://github.com/akshayaaa10/llm-privacy-research}

The repository includes:
\begin{itemize}
    \item Complete experimental framework for memorization detection
    \item Implementation of all four privacy protection methods
    \item Synthetic dataset generation tools
    \item Interactive analysis notebooks with step-by-step tutorials
    \item Comprehensive documentation and usage examples
    \item Docker containers for easy deployment and experimentation
\end{itemize}

\subsection{Reproducibility Guidelines}

To ensure reproducibility, we provide:
\begin{itemize}
    \item Detailed hyperparameter specifications for all experiments
    \item Random seeds and initialization procedures
    \item Hardware requirements and computational resource estimates
    \item Step-by-step replication instructions
    \item Expected output formats and validation procedures
\end{itemize}

\section{Conclusion}

This paper presents the first comprehensive analysis of data memorization risks in fine-tuned Large Language Models and demonstrates effective privacy protection strategies. Our key findings include:

\begin{enumerate}
    \item \textbf{Significant Memorization Risk}: Fine-tuning dramatically increases memorization risks, with leakage rates increasing by an average of 64.2 percentage points across tested models. This represents a critical privacy vulnerability that requires immediate attention in production deployments.
    
    \item \textbf{Effective Privacy Protection}: Our multi-layered approach combining data deduplication, differential privacy, entropy filtering, and content filtering can eliminate data leakage entirely while maintaining 94.7\% of model utility.
    
    \item \textbf{Practical Feasibility}: The computational overhead of comprehensive privacy protection (35
    
    \item \textbf{Actionable Framework}: Our open-source framework enables practitioners to assess and mitigate privacy risks in their own deployments, providing both tools and guidelines for safe LLM deployment.
\end{enumerate}

These findings underscore the critical importance of implementing comprehensive privacy protection measures when fine-tuning LLMs on sensitive data. As LLMs become increasingly prevalent in applications handling personal and proprietary information, the privacy protection strategies presented in this work provide a practical foundation for responsible AI deployment.

The framework we present is not merely theoretical but has been designed for practical implementation in real-world scenarios. Our results demonstrate that it is possible to maintain both strong privacy protection and high model utility, dispelling the notion that privacy and performance are inherently incompatible in LLM applications.

Moving forward, we encourage the research community to build upon our work by extending these methods to larger models, exploring domain-specific optimizations, and developing even more sophisticated privacy protection techniques. The responsible development of AI systems requires continued research into privacy preservation, and we hope our contribution serves as a foundation for future advances in this critical area.

\section*{Acknowledgments}

We thank the open-source community for providing the foundational tools and libraries that made this research possible, particularly the Hugging Face team for their Transformers library, the PyTorch community for their deep learning framework, and the broader machine learning community for their contributions to differential privacy and privacy-preserving machine learning research.

We also acknowledge the computational resources provided by [Institution/Cloud Provider] and the valuable feedback from reviewers and colleagues who helped improve this work.

\end{document}